\documentclass[letterpaper]{article} 
\usepackage[]{aaai23}  
\usepackage{times}  
\usepackage{helvet}  
\usepackage{courier}  
\usepackage[hyphens]{url}  
\usepackage{graphicx} 
\usepackage{natbib}  
\usepackage{caption} 
\usepackage{flushend}
\usepackage{algorithm}
\usepackage{algorithmic}

\usepackage{newfloat}
\usepackage{listings}
\DeclareCaptionStyle{ruled}{labelfont=normalfont,labelsep=colon,strut=off} 
\floatstyle{ruled}
\newfloat{listing}{tb}{lst}{}
\floatname{listing}{Listing}
\title{Pragmatic Constraint on Distributional Semantics}
\author{
    Elizaveta Zhemchuzhina,\textsuperscript{\rm 1}\thanks{This work is an output of a research project implemented as part of the Basic Research Program at the National Research University Higher School of Economics (HSE University). This research was supported in part through computational resources of HPC facilities at HSE University \cite{kostenetskiy2021hpc}.} 
    Nikolai Filippov,\textsuperscript{\rm 1}
    Ivan P. Yamshchikov\textsuperscript{\rm 2, \rm 3}
}
\affiliations{
    \textsuperscript{\rm 1}LEYA Laboratory, Yandex and Higher School of Economics, St. Petersburg, Russia\\
    \textsuperscript{\rm 2}Max Planck Institute for Mathematics in the Sciences, Leipzig, Germany\\
    \textsuperscript{\rm 3}CEMAPRE, University of Lisbon, Portugal\\

    ivan@yamshchikov.info
}

\begin{document}

\maketitle

\begin{abstract}
This paper studies the limits of language models' statistical learning in the context of Zipf's law. First, we demonstrate that Zipf-law token distribution emerges irrespective of the chosen tokenization. Second, we show that Zipf distribution is characterized by two distinct groups of tokens that differ both in terms of their frequency and their semantics. Namely, the tokens that have a one-to-one correspondence with one semantic concept have different statistical properties than those with semantic ambiguity. Finally, we demonstrate how these properties interfere with statistical learning procedures motivated by distributional semantics.

\end{abstract}

\section{Introduction}
\label{intro}

\noindent Various modern Natural Language Processing (NLP) models are designed and trained based on the distributional semantics hypotheses. Namely, that linguistic items used in similar contexts tend to have similar meanings \cite{harris1954distributional}. At the same time, Zipf's law states that given some corpus of natural language utterances, the frequency of any word is inversely proportional to its rank in the frequency table \cite{Zipf1932}. As vocabulary size (or amount of data) grows, this rank-frequency distribution provides a ''heavy tail'' -- a significant amount of linguistic items whose frequency is not enough for distributional semantics to be effective with them. Most modern NLP methods cut this "tail" out of the discussion by introducing the notion of a "token" that could be shorter than the word and restrict the number of tokens in the model's vocabulary. However, the fact that distributional semantics models are trained on the corpora that follow Zipf's law raises a series of interesting  questions such as: 
\begin{itemize}
\item How does a particular tokenization procedure interferes with Zipf's law?
\item Are there distinct differences between frequent "head" and infrequent "tail" tokens?
\item could those differences affect the model's performance and be vital for some aspects of NLP tasks that the model can solve?
\end{itemize}

This paper tries to initiate the discussion of those questions by providing insights into how the empirical fact of the ''heavy tail'' of Zipf's distribution interferes with the methods based on the idea of distributional semantics. 

Since its introduction in the first half of XX century \cite{Zipf1932}, Zipf's law has been extensively studied and applied in various fields: natural languages \cite{Montemurro2001beyond}, random text generation \cite{FerreriCancho2002random}, information theory \cite{Harremoes2006}, population statistics \cite{Reed2002population}, internet traffic statistics \cite{Breslau1999traffic}, infometrics \cite{Egghe2005info}, economics \cite{deWit2005economics}, ecology \cite{Camacho2001ecology}, biology \cite{Gamow1955bio}, information security \cite{Wang2016passwords} just to name a few.

Many researchers studied the significance of Zipf's law for various NLP problems. For example, \cite{yang2013s} suggests that Zipf's law facilitates early language acquisition and uses Zipf's law to distinguish human language usage from bioacoustics of other species, such as chimps. \cite{zhang2008exploring} discovers that the distribution of lexical tokens in Java source code follows the Zipf's law. \cite{takahashi2017neural} argue that a signature of efficient representations is that frequency distributions follow power laws.  \cite{pimentel2021non} demonstrate that natural codes are closer to not being optimized (in the Zipfian sense) than to being maximally compressed. \cite{Cristelli2012} find that many natural systems do not show true power law behavior because they are incomplete or inconsistent with the conditions under which one might expect power laws emergence.  \cite{ferrer2018origins} show that a single assumption on the joint probability of a word and a meaning suffices to infer Zipf's meaning-frequency law and argue that this assumption can be justified as the outcome of a biased random walk in the process of mental exploration. Finally, \cite{nikkarinen2021modeling} argue that any approach that assigns zero probability to any out-of-vocabulary word form produces negatively biased probabilities for any out-of-vocabulary word, while positively biased probabilities to in-corpus words. The authors make a compelling argument in favor of properly modeling the unigram distribution and claim it should be a central task in NLP.

In this paper, we study the rank-frequency distribution behavior for sets of tokens with different maximum token lengths (derived from vocabulary size and tokenization algorithm) and show that this distribution is well approximated with the superposition of two Zipf's laws for two distinct, coherent sets of tokens. We infer that subsets of tokens sampled from the head and tail of such distributions have differences both in terms of statistics and semantics. Thus we introduce the terms ''pragma'' and ''idea'' to distinguish between the two subsets of tokens. We believe that conceptual understanding of these two token categories is fundamental for assessing the limitations of current models based on the so-called distributional hypothesis \cite{harris1954distributional}.

\section{Experiments}

Here we present a series of experiments that provides insights into statistic and semantic properties of various tokens provided by the chosen tokenization. 

\label{experiments}

\subsection{Zipf's Law for Different Tokenizations}
\label{experiments:tokenizations}

We experiment with three different tokenization algorithms: Byte Pair Encoding or BPE \cite{Gage1994bpe}, WordPiece \cite{wu2016wordpiece}, and Unigram \cite{Kudo2018unigram}. We find no major differences regarding token frequency vs. token rank behavior. Up to some vocabulary size limit, the rank-frequency distribution follows Zipf's law regardless of the chosen tokenization algorithm.

\begin{figure}
\centering
  \includegraphics[width=.8\linewidth]{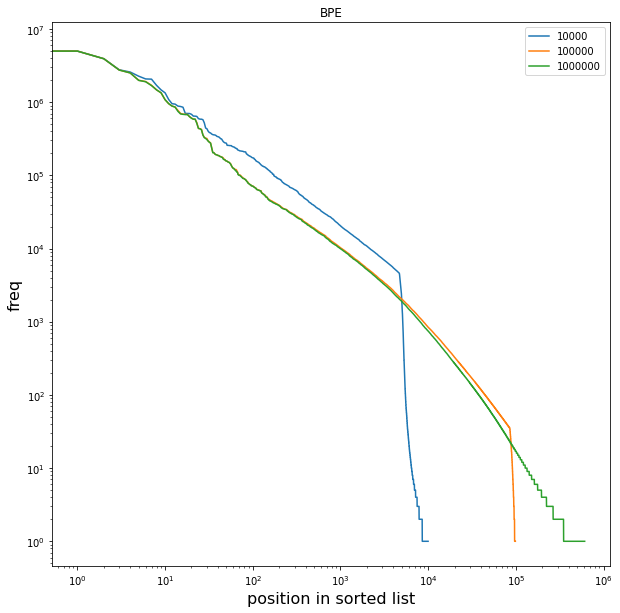}
  \includegraphics[width=.45\linewidth]{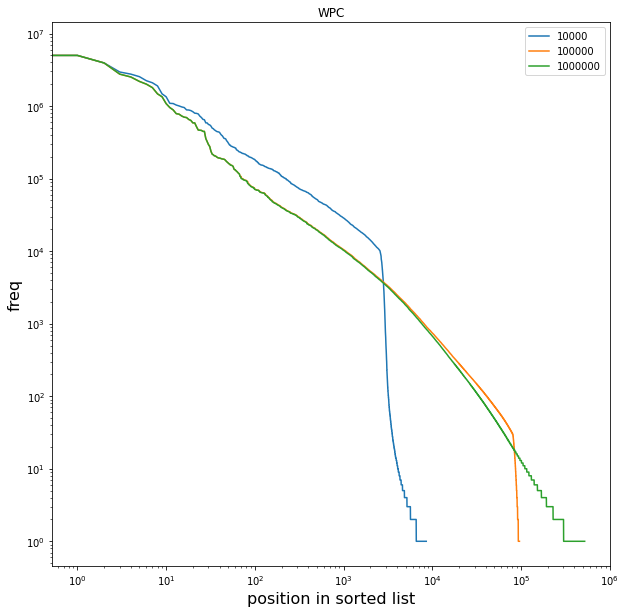}
  \includegraphics[width=.45\linewidth]{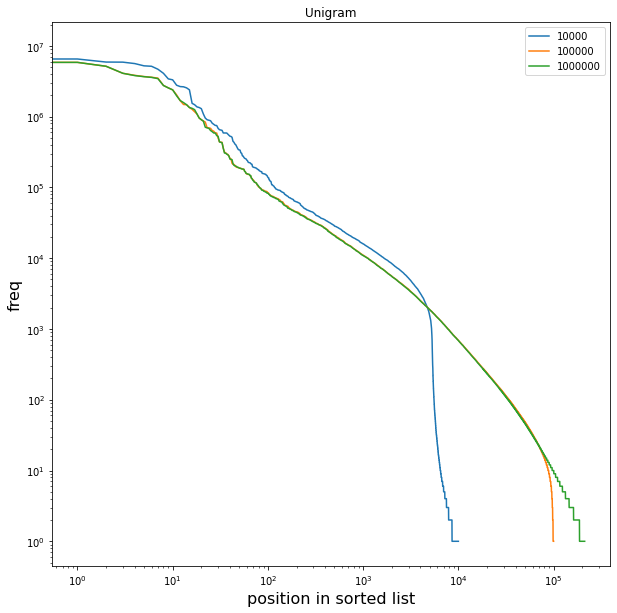}
\caption{Rank-Frequency distribution for BPE, WordPiece and Unigram tokenization algorithms, vocabulary sizes 10~000, 100~000, 1~000~000}
\label{fig:zipf}
\end{figure}

Figure~\ref{fig:zipf} shows the rank-frequency distributions of tokens for Wikitext-103\footnote{Wikitext-103 dataset was used for all experiments reported in the paper. https://huggingface.co/datasets/wikitext}. dataset with BPE, WordPiece, and Unigram tokenization, respectively, with three different vocabulary sizes. All of them demonstrate Zipf-like behavior. We found BPE to be the most illustrative algorithm for this paper due to the nature of the algorithm. BPE would produce longer tokens as vocabulary size increases since BPE keeps adding the most frequent pair of existing tokens to the vocabulary as a new token. This makes it straightforward: a bigger vocabulary size leads to a bigger maximum token length. From now on, if the tokenization algorithm is not explicitly specified, BPE is used.

\subsection{Closer Look at Zipf's Law}
\label{experiments:closer}

We run a series of experiments with different vocabulary sizes, from a few thousand (which gives tokenization on the symbol, N-gram, and word levels) to several million (which inevitably adds tokens consisting of several words and/or whole phrases to the vocabulary). We found that, with smaller vocabulary sizes, the rank-frequency distribution tends to follow Zipf's law even on the level of sub-word tokens. As the vocabulary size grows further to include tokens consisting of several words and/or whole phrases in the vocabulary, the rank-frequency diagram does not anymore follow the ''pure'' power law but rather resembles the superposition of two Zipf's laws, thus following the ''coherence'' concept introduced by Matthieu Cristelli, Michael Batty and Luciano Pietronero \cite{Cristelli2012}. Once the vocabulary size reaches some point, there appears a threshold near which the rank-frequency diagram experiences a phase transition. The parameter of Zipf's approximating rank-frequency distribution shifts. We presume that such behavior is caused by the fact that, in the case of larger vocabularies, we deal with two subsets of tokens, each subset being coherent in itself. Figure~\ref{fig:bpe1m} in Appendix~\ref{appendix} represents the rank-frequency distribution for vocabulary size 1 million, illustrating the ''bend'' of Zipfian distribution. This figure represents BPE tokenization algorithm, but we observe the same behavior for other tokenization algorithms as well. For further experimental results, we address the reader to the Appendix~\ref{appendix}.

\subsection{Semantics and Zipf's law}
\label{experiments:semantics}

As seen from the experiment results (see diagrams in Appendix~\ref{appendix}), the distribution starts behaving as a superposition of two Zipf's laws at the point where token length starts exceeding some threshold. We presume that this threshold is determined by semantics: the distribution behavior at the head part of such diagrams differs from that at the tail part. The hypothesis is that the head part mostly consists of shorter tokens with possible semantic variations, whereas the tail part mostly consists of longer tokens connected to one specific semantic field. Figure~\ref{fig:len1m} in Appendix~\ref{appendix} illustrates the token lengths distribution. Both shorter multi-meaning and longer single-meaning tokens are coherent and demonstrate distribution under Zipf's law if considered separately. This is visible on distribution diagrams with smaller vocabulary sizes that, presumably, consist mostly of shorter tokens (see example at Figure~\ref{fig:bpe5k_all} in Appendix~\ref{appendix}). Together these two subsets do not demonstrate pure Zipfian behavior anymore. A ''heavy tail'' of such distribution is clearly visible at Figure~\ref{fig:bpe1m} in Appendix~\ref{appendix}. Figure~\ref{fig:len1m} in Appendix~\ref{appendix} shows that the tail mostly consists of longer tokens which, as shown below, tend to have one meaning rather than many.

To illustrate qualitative differences between these two distributions, we carried out an additional experiment with two subsets of tokens: one from the head of the one-million-tokens vocabulary distribution and another one from the tail. As expected, the ''head'' subset consisted of shorter tokens, primarily words, and the ''tail'' subset consisted of longer tokens, mainly phrases and parts of sentences. Since there are no long tokens neither in the head nor in the middle of the distribution (see Figure~\ref{fig:len1m} in Appendix~\ref{appendix}), and because of the nature of the experiment (we were mainly interested in the audience's perception of shorter tokens from the head and longer tokens from the tail), we filtered out a few short tokens that might occur in the tail, and left the tail part with long tokens only. We conducted a poll regarding the shuffled sequence of tokens from the two subsets among a group of professionals in linguistics to find out the difference between these subsets based on the opinion of people with relevant professional backgrounds. We asked the following questions regarding every single token $X$ in the set:

\begin{enumerate}
    \item Can you reformulate $X$?
    \item How many meanings does $X$ have depending on context?
    \item Can you place $X$ into context?
\end{enumerate}

\begin{figure}
\includegraphics[width=0.9\linewidth]{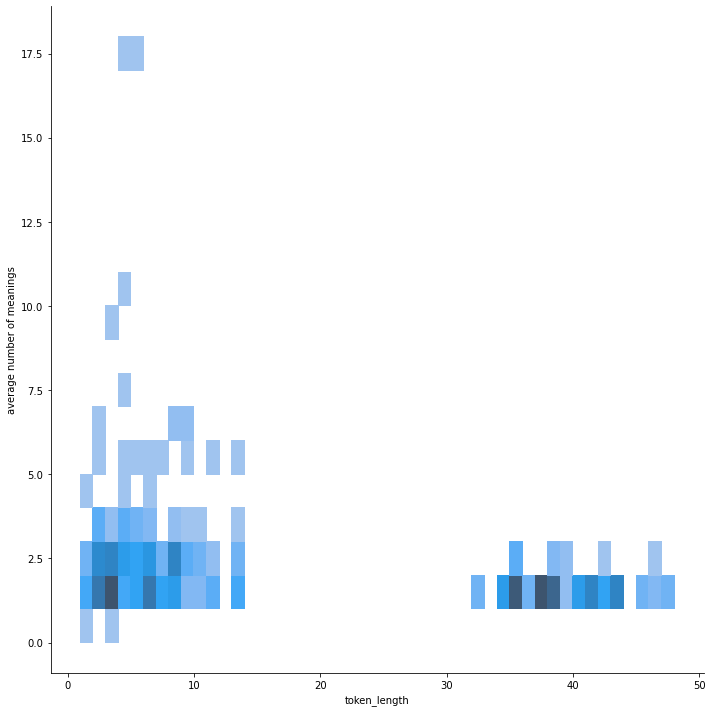}
\caption{Poll results: token length vs. average number of meanings (heatmap) listed by a linguist} \label{fig:len_meaning}
\end{figure}

Figure~\ref{fig:len_meaning} illustrates one of the poll results: shorter tokens from head part of distribution are characterized with semantic ambiguity and more often can have several meanings depending on context. Longer tokens from tail part of distribution tend to have one or two specific meanings.

Figure~\ref{fig:len_levenstein} in Appendix~\ref{appendix} illustrates that the shorter, semantically ambiguous tokens can be easily placed into different contexts, while longer, single-meaning tokens represent context themselves: attempt to place them in context results in changing a few shorter tokens they consist of, leaving the rest unchanged. The difference is clearly visible by the normalized Levenstein distance between the original token and the token in context. It is also worth mentioning that the normalized Levenstein distance distribution in the left part of the diagram visually resembles Zipf's law as well -- this may be a topic for a closer look and more detailed study with a much bigger audience.

The results were clustered in accordance with the subset each token belonged to and show that based on the opinion of people with professional linguistic backgrounds:
\begin{itemize}
    \item the ''head'' tokens can be easily replaced with synonyms, homonyms, or phrases with synonymous or homonymous meaning, the full content of the token is changed in the course of restatement, along with (possibly) meaning;
    \item the ''tail'' tokens are difficult to reformulate, mainly through changing one or more shorter tokens they consist of but leaving the rest unchanged. Regardless of the restatement, the whole token's meaning does not change;
    \item ''head'' tokens often have several meanings while ''tail'' tokens mostly have one or two meanings;
    \item ''head'' tokens can be easily placed into context and may have different meanings depending on the context;
    \item placing the ''tail'' token into context is not that easy and is done by placing a few shorter tokens into the context of the longer token rather than vice versa, while the whole long token's meaning remains unchanged.
\end{itemize}

These results illustrate that ''head'' and ''tail'' tokens differ both in terms of semantics and in terms of statistics. ''Head'' tokens have semantic ambiguity, can be easily placed in different contexts with different meanings. ''Tail'' tokens have a one-to-one correspondence with the specific semantic concept. They are not easy to place in a new context other than the context they define themselves. The tail part of rank-frequency distribution, mostly consisting of such tokens, demonstrates behavior inherent to the rank-frequency distribution of distinct, coherent sets of tokens. Both ''head'' and ''tail'' subsets are coherent in terms of \cite{Cristelli2012}, and the distribution of each of them seems to follow Zipf's law. In contrast, the distribution of their union looks more like a superposition of two Zipf's laws associated with two different coherent sets of tokens.

\section{Pragma and Idea}
\label{pragma_idea}

We suggest to introduce the following term that characterize different types of tokens:

\textit{Atom} -- the smallest element of a discrete sequence that cannot be divided further into smaller parts. In written language, the single symbols are considered atoms. The initial set of atoms is believed to be an exhaustive list limited in size.

\textit{Pragma} -- a part of a discrete sequence that consists of atoms and represents an integral part of an idea. Pragma has semantic ambiguity and may have different meanings depending on context. In written language, pragmas are represented by symbol n-grams, words, and word n-grams.

\textit{Idea} -- a part of a discrete sequence that consists of pragmas and has a one-to-one correspondence to some specific concept (a semantic concept in the case of language). In written language, ideas can be represented by semantically distinct sentences, set phrases, and colloquialisms having specific meanings.

We do not mention atoms in this paper, but we introduce them here for integrity. We believe such an atom-pragma-idea granularity structure helps to understand and describe some concepts related to both statistical behavior and semantics of specific sets of tokens. We also believe that such a three-layer granularity structure may find its application in discrete sequence processing studies related to many fields other than natural language: chemistry, biology, music, etc.

\section{Discussion}
\label{discussion}

Following the experiment results and findings of the experiments, we consider some topics as subjects for further investigation and discussion.

Language models tend to use vocabulary sizes smaller than the ones we used in our experiments. Therefore, they represent ideas as sequences of pragmas rather than forming a distinct, coherent subset of tokens out of ideas. This makes it difficult to register the ''phase shift'' where the longer sequence of pragmas becomes an idea: it starts to demonstrate one-to-one correspondence to specific semantic concept.

We believe this might be one of the reasons why language models show poor performance when dealing with longer sequences, in contrast to humans who do not have such problems. We suggest addressing this phenomenon as {\em Pragmatic Constraint} — the capability of statistical learning to operate within pragmatic tokenization that does not leverage the reduction of semantic ambiguity characteristic for the idea-based tokens.

\section{Conclusion}
\label{conclusion}

This paper studies the statistical and semantic properties of tokens. Both statistical and semantic differences were found for two distinct subsets of tokens that we called ''pragma'' and ''idea''. Pragmas have semantic ambiguity, while ideas have a one-to-one correspondence to the specific semantic concept. Statistically, the set containing both of these subsets does not demonstrate pure Zipfian behavior but rather behaves like a superposition of two Zipf's laws. Results of a poll conducted among a small group of linguists are in line with these findings. This may impose limitations on machine learning methods and techniques based on statistical learning and assumptions about the Zipfian nature of sets of tokens. Such limitations and their impacts are subject to further study.

\section*{Limitations}

Wikitext-103 dataset was used for this research. We believe the results are reproducible with other datasets in languages other than English, but this has not been proved yet.

For results to be reproducible, the dataset shall be big enough to have at least a million tokens in vocabulary.

The poll was performed among a small group of five professionals since the primary goal was not to gather statistics but rather to get the opinions of people with strong professional background in linguistics. For statistically significant results, a broader audience is required.

GPU is preferable to achieve results within a reasonable time.

\bibliography{custom.bib}

\clearpage

\appendix

\section{Appendix}
\label{appendix}

Below are figures representing experiment results referred to in the article.

\begin{figure}[H]
\includegraphics[width=0.48\textwidth]{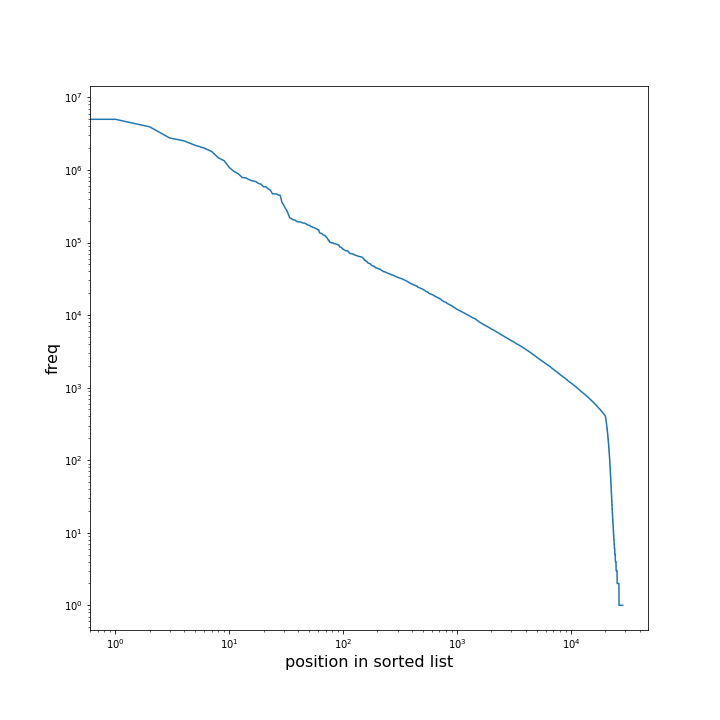}
\caption{Rank-Frequency distribution: WordPiece tokenization, vocabulary size 30~000 (rare tokens excluded)} \label{fig:wpc30k}
\end{figure}

\begin{figure}[H]
\includegraphics[width=0.48\textwidth]{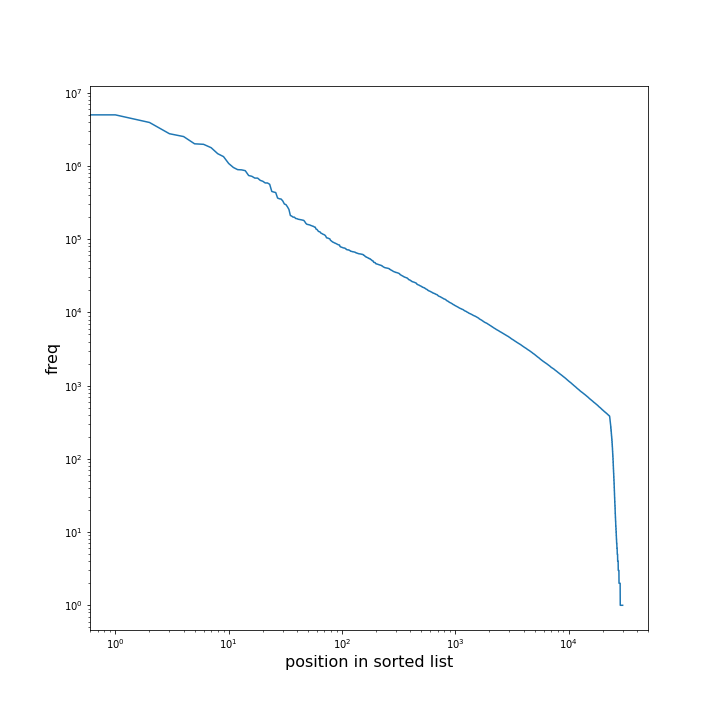}
\caption{Rank-Frequency distribution: BPE tokenization, vocabulary size 30~000 (rare tokens excluded)} \label{fig:bpe30k}
\end{figure}

Figures~\ref{fig:wpc30k}--\ref{fig:bpe30k} illustrate Zipf's law behavior for different tokenization algorithms.

\begin{figure}[H]
\includegraphics[width=0.48\textwidth]{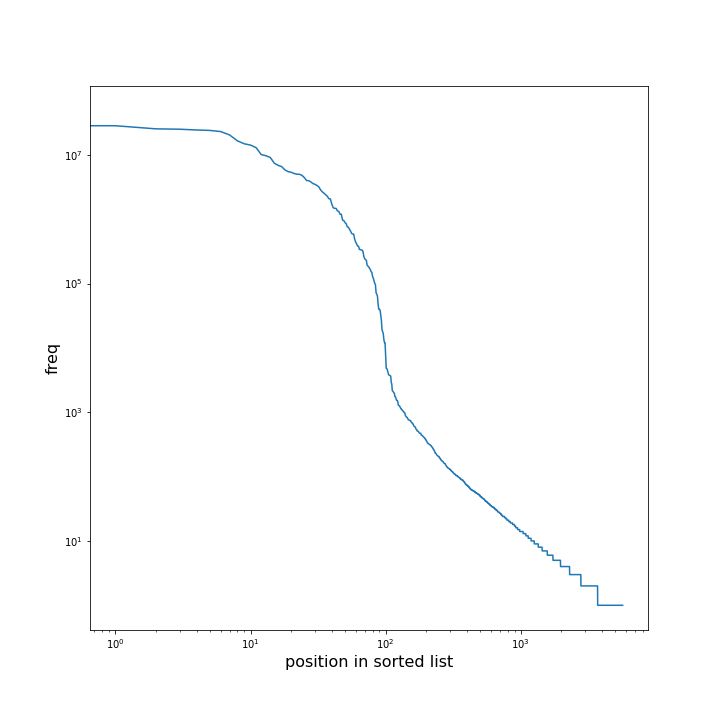}
\caption{Rank-Frequency distribution: vocabulary size 5~000} \label{fig:bpe5k_all}
\end{figure}

\begin{figure}[H]
\includegraphics[width=0.48\textwidth]{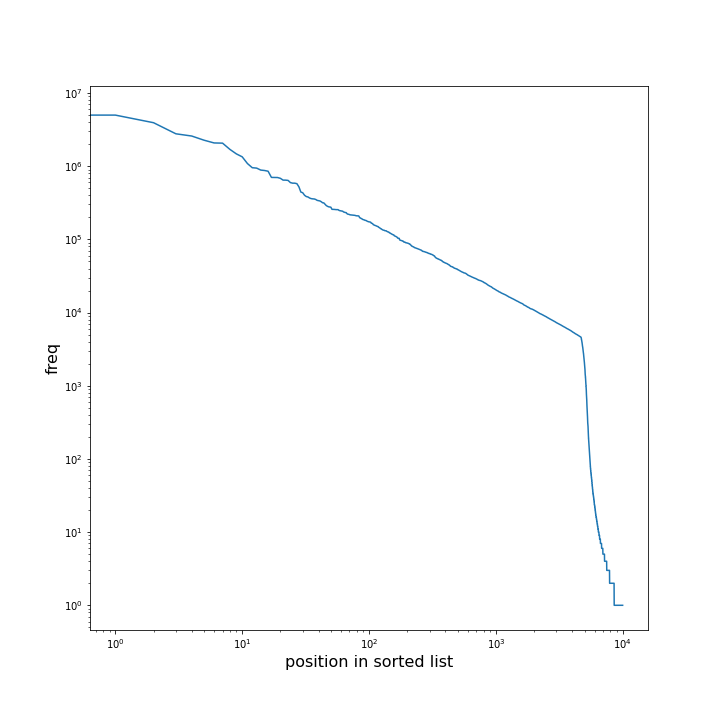}
\caption{Rank-Frequency distribution: vocabulary size 10~000} \label{fig:bpe10k_all}
\end{figure}

\begin{figure}[H]
\includegraphics[width=0.48\textwidth]{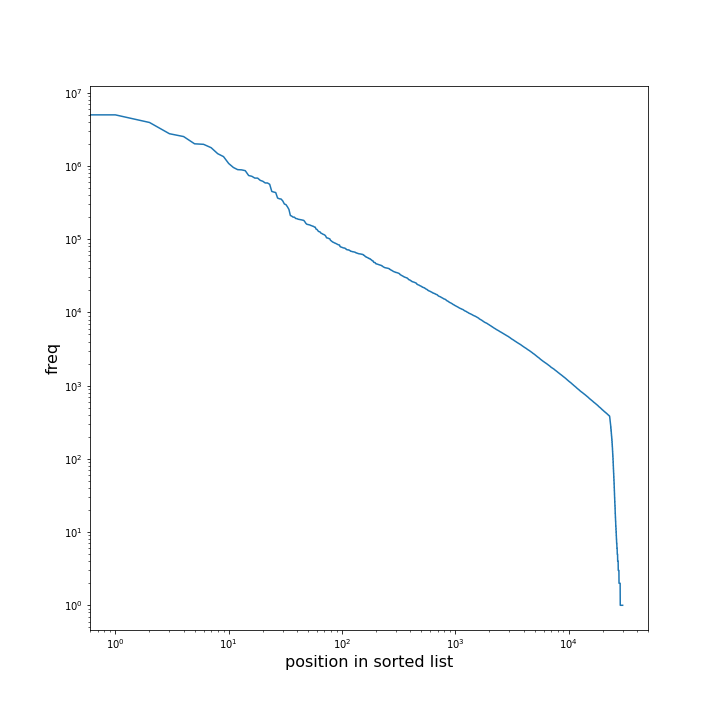}
\caption{Rank-Frequency distribution: vocabulary size 30~000} \label{fig:bpe30k_all}
\end{figure}

\begin{figure}[H]
\includegraphics[width=0.48\textwidth]{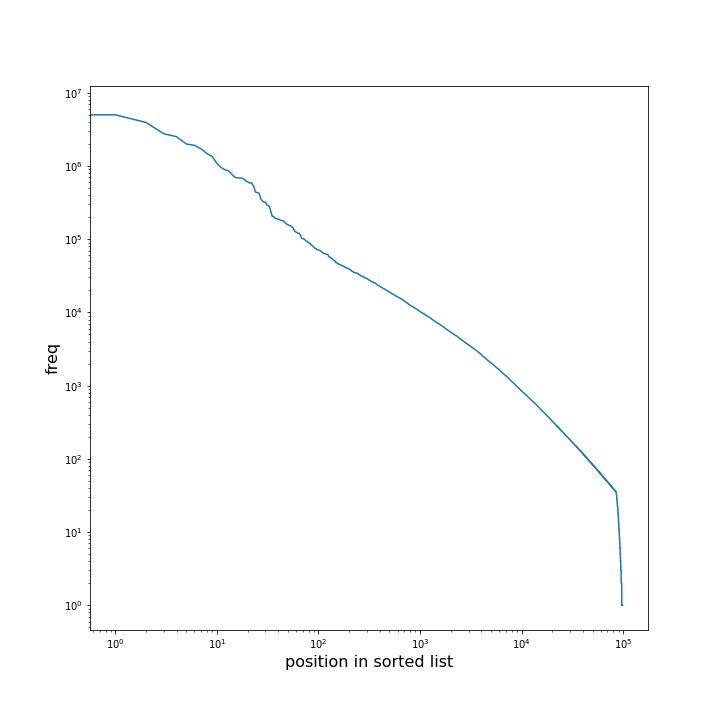}
\caption{Rank-Frequency distribution: vocabulary size 100~000} \label{fig:bpe100k_all}
\end{figure}

Figures~\ref{fig:bpe5k_all}--\ref{fig:bpe100k_all} illustrate Zipf's law behavior for different vocabulary sizes and, hence, different maximum token lengths.

\begin{figure}[H]
\centering
  \includegraphics[width=.78\linewidth]{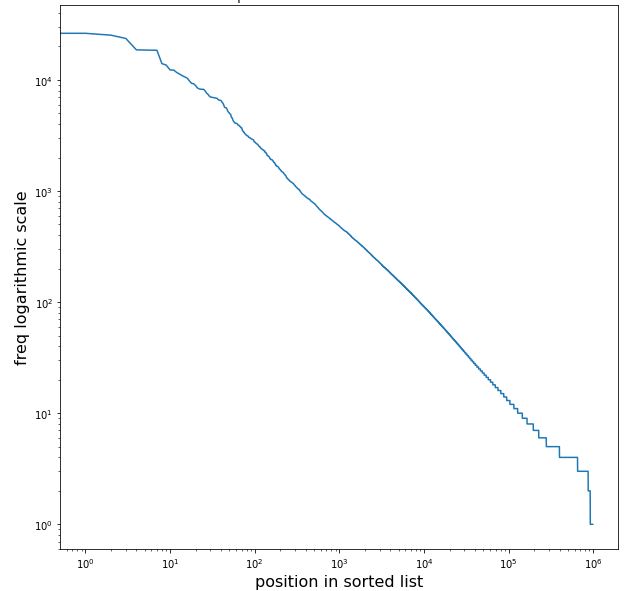}
\caption{Rank-Frequency distribution and token lengths distribution: BPE tokenization algorithm, vocabulary size 1~000~000}
\label{fig:bpe1m}
\end{figure}

\begin{figure}[H]
\includegraphics[width=0.45\textwidth]{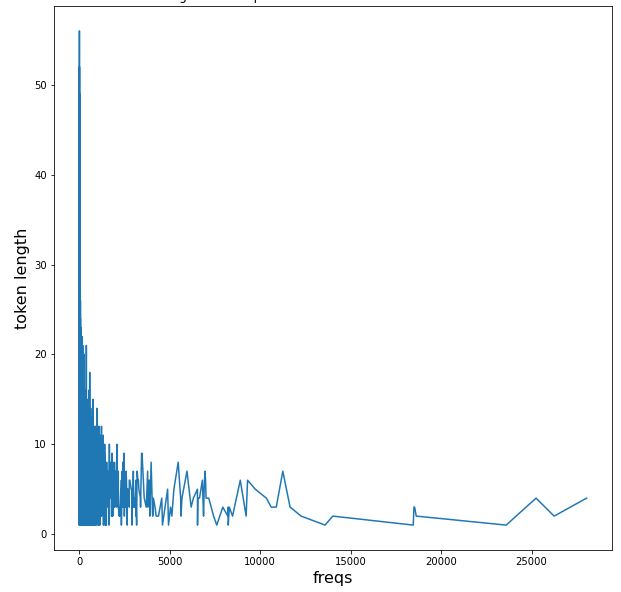}
\caption{Token Lengths frequency: vocabulary size 1 million, logarithmic scale} \label{fig:len1m}
\end{figure}

Figures~\ref{fig:bpe1m}--\ref{fig:len1m} represents token lengths distribution for vocabulary size 1~000~000.

\begin{figure}[H]
\includegraphics[width=0.96\linewidth]{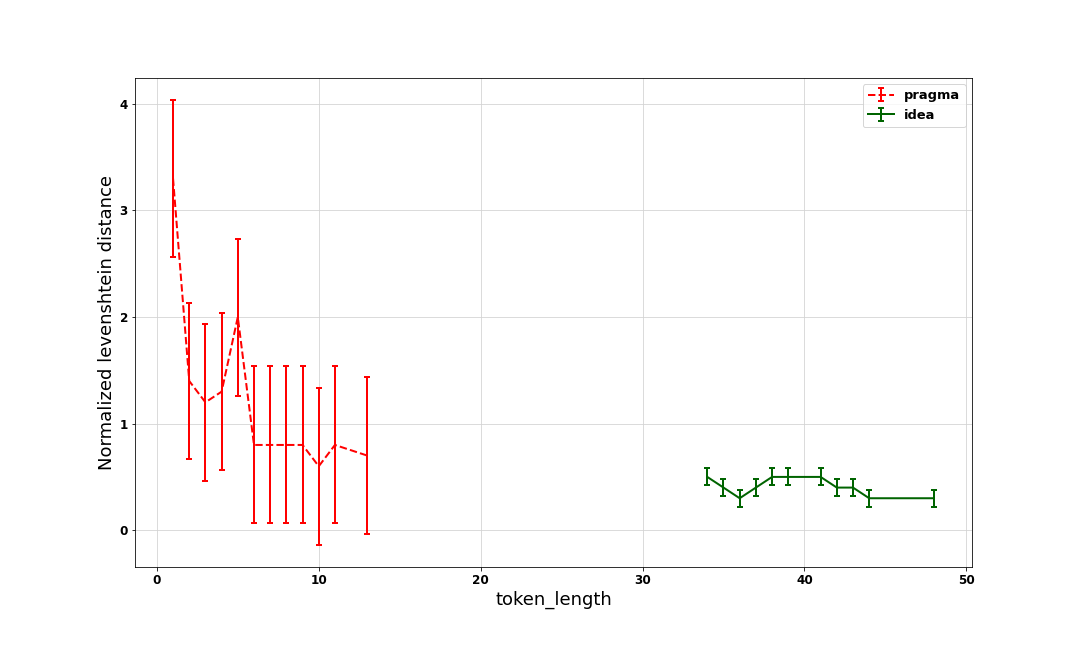}
\caption{Poll results: token length vs. Levenstein distance between original token and token after restatement} \label{fig:len_restate}
\end{figure}

\begin{figure}[H]
\includegraphics[width=0.96\linewidth]{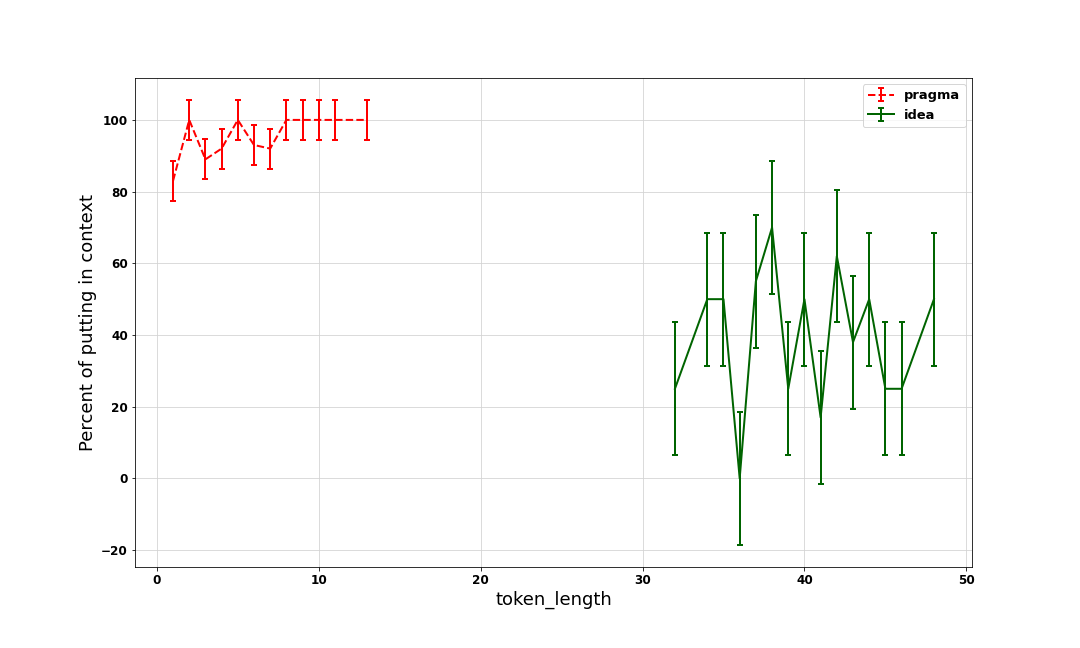}
\caption{Poll results: token length vs. percentage of token placed in context} \label{fig:len_context}
\end{figure}

\begin{figure}[H]
\centering
  \includegraphics[width=.96\linewidth]{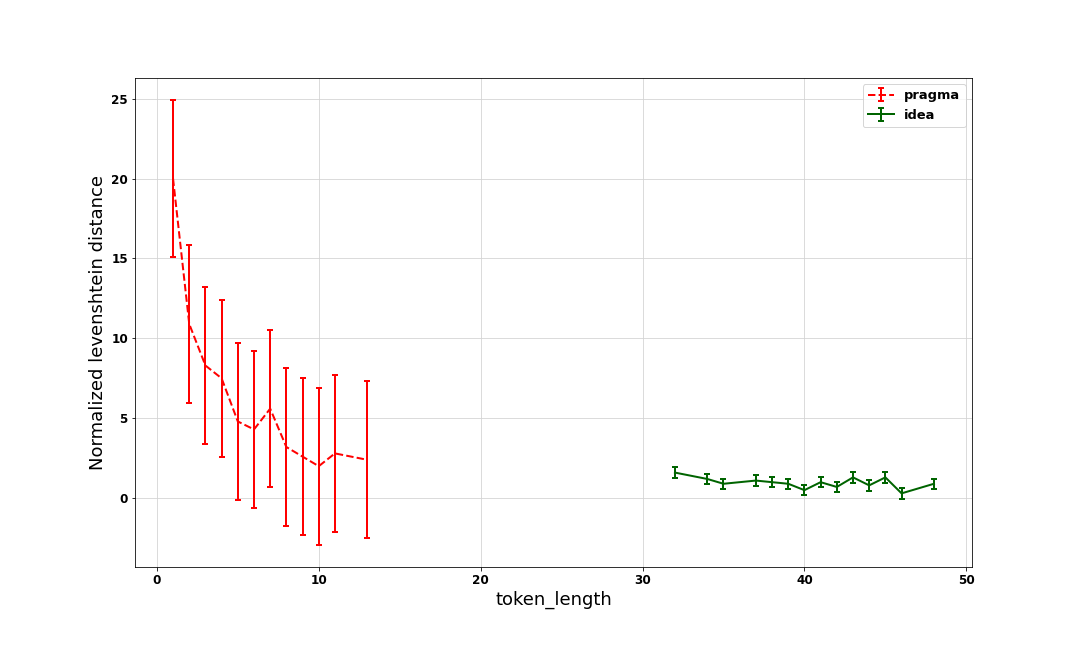}
\caption{Poll results: token length vs. normalized Levenstein distance between original token and token placed in context. "Head" tokens can be placed in various contexts that differ significantly. ''Tail'' tokens tend to be embedded in the similar contexts that are comparable with the size of the token itself.}
\label{fig:len_levenstein}
\end{figure}

Figures~\ref{fig:len_restate}--\ref{fig:len_levenstein} illustrate poll results: average Levenstein distance between original token and token after restatement for different token lengths, percentage of audience that could place the token in context for different token lengths, and normalized Levenstein distance between original token placed in context for different token lengths.

\end{document}